\documentclass[runningheads]{llncs}
\usepackage{graphicx}
\usepackage{amsfonts}
\usepackage{hyperref}
\begin{document}
\title{QuIIL at T3 challenge: Towards Automation in Life-Saving Intervention Procedures from First-Person View}
\author{Trinh T.L. Vuong \inst{1*} \and
Doanh C. Bui \inst{1*} \and
Jin Tae Kwak \inst{1}}
\authorrunning{Vuong et al.}
\titlerunning{QuIIL at T3 challenge}

\institute{School of Electrical Engineering, Korea University}
\def\thefootnote{*}\footnotetext{First authors contributed equally.}
\maketitle
\begin{abstract}
In this paper, we present our solutions for a spectrum of automation tasks in life-saving intervention procedures within the Trauma THOMPSON (T3) Challenge, encompassing action recognition, action anticipation, and Visual Question Answering (VQA). For action recognition and anticipation, we propose a pre-processing strategy that samples and stitches multiple inputs into a single image and then incorporates momentum- and attention-based knowledge distillation to improve the performance of the two tasks. For training, we present an action dictionary-guided design, which consistently yields the most favorable results across our experiments. In the realm of VQA, we leverage object-level features and deploy co-attention networks to train both object and question features. Notably, we introduce a novel frame-question cross-attention mechanism at the network's core for enhanced performance. Our solutions achieve the \(2^{nd}\) rank in action recognition and anticipation tasks and \(1^{st}\) rank in the VQA task. 
The source code is available at \url{https://github.com/QuIIL/QuIIL_thompson_solution}.

\end{abstract}

\section{Introduction}

Scene understanding and analysis are essential tasks in AI systems that provide insights into a scene and allow for the prediction of appropriate actions. This is, in particular, critical in developing medical computer-assisted systems such as remote instruction systems in uncontrolled and austere environments, especially for life-saving procedures \cite{kirkpatrick2017damage}. Such capabilities can provide valuable support to both first responders and individuals lacking specialized training in the medical scenario. For instance, in a situation where a patient is experiencing bleeding caused by a blast injury, the necessary response could involve the application of a tourniquet. In this particular scenario, it is necessary to develop an AI-based computer-assisted system from a first-person perspective. This is fundamentally crucial since it precisely emulates the visual perception of an individual possessing expertise in first aid, thereby offering an unequivocally clear and comprehensive account of the events transpiring around an injured person. However, the resources for the mentioned problem are still restricted. While the field of first-person view scene analysis for daily-living activities has seen a proliferation of datasets \cite{pirsiavash2012detecting,calway2015discovering,li2018eye,damen2020epic}, the availability of such invaluable resources in the context of medical applications remains limited. Leveraging advanced methods and transferring pre-trained weights from other domains for medical procedure prediction is one of the potential solutions to overcome dataset limitations. 

In this manuscript, we present a comprehensive description of our approaches to address three tasks in the Trauma THOMPSON (T3) challenge, including action recognition, anticipation, and visual question answering. To provide a succinct overview, the key methods can be outlined as follows:

\begin{itemize}
    \item In the action recognition and anticipation tasks, we introduce a pre-processing strategy and action dictionary-guided (ADG) learning design. We also exploit the pre-trained model, which is adapted to life-saving procedures, through an advanced knowledge distillation method, i.e., Momentum contrastive learning with Multi-head Attention (MoMA) \cite{vuong2023moma}. This ADG-learning strategy results in our best-performing experiments compared to single-task and multi-task learning.

    \item In the visual question answering task, we leverage the object features extracted by the pre-trained VinVL model \cite{zhang2021vinvl}, which excels at capturing first-view camera context. To further enhance the approach, we introduce frame-question cross-attention based on the MCAN baseline network \cite{yu2019deep}. This integration effectively identifies crucial object features, leading to more accurate answers to questions. We show that the proposed solution obtained the best performance among our three experiments.
\end{itemize}

The rest of this paper is structured as follows: Section \ref{sec:method} provides a comprehensive overview of our solutions for the three tasks; Section \ref{sec:results} reports the results and discussions; Section \ref{sec:conclusion} summarizes our report for the T3 challenge.

\section{Methodology}\label{sec:method}
\subsection{Track 1 - Task 1 \& 2: Action Recognition \& Anticipation }
\begin{figure}[http]
\centerline{\includegraphics[width=1\linewidth]{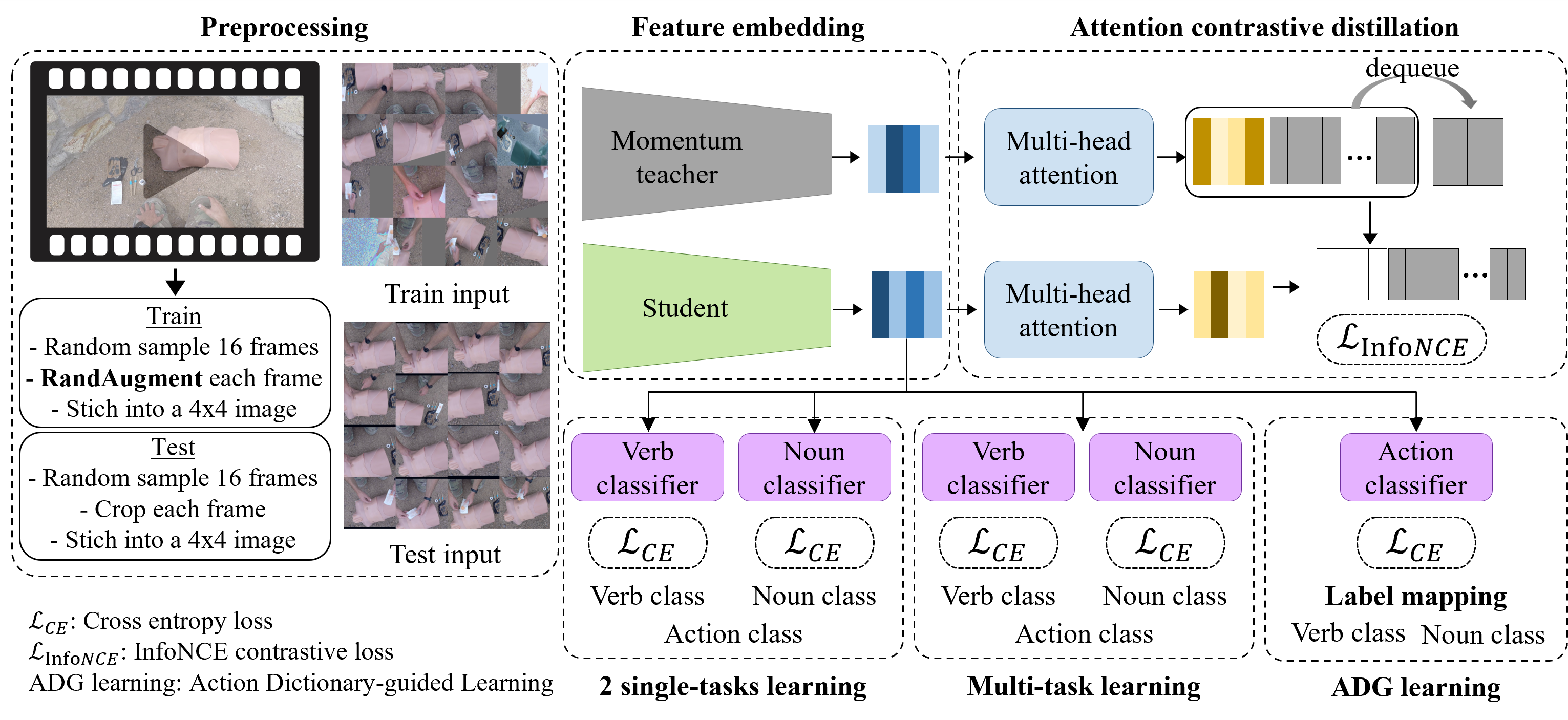}}
\caption{Overview of ADG: Action Dictionary-guided learning model for action recognition and action anticipation task.}
\label{cls_model}
\end{figure}

Overview of the proposed method, as shown in Figure \ref{cls_model} includes four components: the pre-processing module, feature embedding, attention contrastive distillation, and action dictionary-guided (ADG) classification head. 

\subsubsection{Problem definition.} For two tasks involving action recognition and anticipation, given a sequence of \(N^f\) frames \(\mathbf{F}_i = \{f^{(i)}_k\}_{k=1}^{N^f}\), the objective is twofold: to recognize the current action \(\mathbf{\hat y_i}\) and to predict the subsequent action \(\mathbf{\hat y_{i+1}}\) based on the next sequence of frames \(\mathbf{F}_{i+1}\). Notably, action recognition (\(\mathbf{y}_i\)) relies on two predicted components: the verb (\(\hat v_i\)) and the noun (\(\hat n_i\)), a structure similarly mirrored in \(\mathbf{y}_{i+1}\).

\subsubsection{Action Dictionary-guided Learning}

\begin{table}[h]
\centering
\caption{Illustrative examples of verb-noun pairings and corresponding actions for action recognition and anticipation}
\setlength{\tabcolsep}{10pt} % Default value: 6pt
\resizebox{1\textwidth}{!}{\begin{tabular}{cccccc}
\hline
Verb class &	Verb &	Noun class &	Noun &	Action class &	Action \\
\hline
4 &	attach &	36 &	syringe &	0 &	attach syringe \\
26 &	prep &	28 &	site &	1 &	prep site \\
33 &	take &	36 &	kelly &	2 &	take kelly \\
$\cdots$ & $\cdots$ & $\cdots$ & $\cdots$ &$ \cdots$ & $\cdots$ \\
4 &	attach &	0 &	ambubag &	113 &	attach ambubag \\
\hline
\multicolumn{2}{c}{\# Verb Classes = 41} & \multicolumn{2}{c}{\# Noun Classes = 45} & \multicolumn{2}{c}{\# Action Classes = 114} \\

\hline
\end{tabular}}
\label{tab:AGD}
\end{table}

This section outlines our Action Dictionary-guided (ADG) design for the action recognition and anticipation tasks. We begin with two sets: a set of verb labels, denoted as $V = \{v_1, v_2, \ldots, v_k\}$, and a set of noun labels, denoted as $N = \{n_1, n_2, \ldots, n_h\}$. From these sets, we create a dictionary of unique action labels, represented as $A = \{(v_1, n_1), (v_2, n_2), \ldots, (v_k, n_h)\}$.

In our ADG design, we construct an extended action label dictionary, denoted as $\hat{A} = \{a_1: (v_1, n_1), a_{2}: (v_2, n_2), \ldots, a_g: (v_k, n_h)\}$. The model's training involves learning these action labels from $\hat{A}$. During inference, the model maps these action labels back to their respective verb and noun classes. In our dataset, we work with \(41\) verb classes ($k=41$), \(45\) noun classes ($h=45$), and a total of \(114\) unique action classes ($g=114$). Illustrative examples of verb-noun pairings and corresponding actions for action recognition and anticipation are shown in Table \ref{tab:AGD}.

To assess the effectiveness of our learning design, we also compare it with a single-task classification approach, where the model separately learns verb and noun classes. In this setup, two models for verb and noun classification are trained independently. However, it is important to note that verb and noun labels often exhibit relationships within the same video context. Learning them separately tends to yield suboptimal results. To address this, we adopt a multi-task learning approach, allowing the model to share valuable features between the tasks and consequently improve performance.

In our multi-task model, we employ a shared encoder for both verb and noun classification tasks, each with its own dedicated classification head. In contrast, combination task learning requires only a single action classifier head. 

In all classification tasks, including ADG, single-task, and multi-task learning, we optimize using the cross-entropy loss, denoted as $\mathcal{L}_{CE}$. It is worth highlighting that the single-task learning approach necessitates training two separate models to obtain action predictions. In multi-task learning, one encoder and two classification heads must be trained. In contrast, our ADG learning approach is the most parameter-efficient option, as it requires only one encoder and one classification head to be trained.

\subsubsection{Video-to-image prepossessing.} In the training phase, given a sequence of frames \(\mathbf{F}_i\) including \(N^f\) frames \(f^{(i)}_k \in \mathbb{R}^{3 \times H \times W}\), we first randomly select \(N^{f'} < N^f\) frames (specially, \(N^{f'} = 16\)). Each frame is resized to one-fourth (1/4) of its original size, then randomly cropped to $224\times224$, then applied the RandAugment augmentation strategy for each frame. Then, \(N^{f'}\) selected augmented frames are stiched as a \(\sqrt{N^{f'}} \times \sqrt{N^{f'}}\) window \(\mathbf{F}_i' \in \mathbb{R}^{ 3 \times (\sqrt{N^{f'}} \times H) \times ( \sqrt{N^{f'}} \times W)}\). Here in, \(\mathbf{F}_i'\) are treated as a single input image when processing \(\mathbf{F}_i'\) through the backbone network.
In the testing phase, a similar strategy from training is applied to each testing video \(\mathbf{F}_i\) obtain \(\mathbf{F}_i'\). However, for each frame, we start by center cropping a region of $(224 \times 1.3) \times (224 \times 1.3)$ and subsequently perform a random crop of $(224) \times (224)$ within this area. No RandAugment is applied during this process. This repeat is repeated $n=30$ times for each input video.

\subsubsection{Image-MoMA.} 

We came to the realization that starting the CNN backbone network training from scratch did not yield satisfactory performance. Consequently, we opted to leverage the insights from the MoMA study \cite{vuong2023moma}, which empowered our model to effectively inherit knowledge from a pre-trained CNN backbone, originally trained on the ImageNet dataset, thereby enhancing its performance on the challenging tasks at hand. First, we convert the input video into $4 \times 4$ frames images. We refer to it as Image-MoMA, as we flatten a video into a single image to input it into image-based deep learning models, which is different from other video processing methods in  Video-ViT\cite{wang2023masked}.
There are two types of networks in the MoMA setting: the momentum teacher \(f^{(T)}\) that was trained on the ImageNet dataset, and the student model \(f^{(S)}\) that will be trained on a target-domain dataset, which is provided by the T3 challenge organizer. \(f^{(T)}\) network is frozen during the training process, and only weights of \(f^{(S)}\) are updated.

The MoMA framework actively influences the relationships between samples within a batch during each iteration. In the \(k^{th}\) iteration, two networks process a batch denoted as \(\mathbf{B}_k = \{\mathbf{F}^{(k)}_i\}_{k=1}^{bs}\). Subsequently, \(f^{(T)}\) and \(f^{(S)}\) generate two distinct sets of embedding vectors, \(\mathbf{E}^T_k\) and \(\mathbf{E}^S_k\), respectively. Following this, a multi-head self-attention mechanism is applied to each of these sets, effectively recalibrating the importance of the embedding vectors contained within them. There exists a memory bank \(\mathbf{queue}\) with a length of \(L_N\), designed to store a set of negative samples \(\mathbf{E}^T_k\), denoted as \(\mathbf{queue^T} = \{\mathbf{E}^T_l\}_{l=1}^{L_N}\). Subsequently, contrastive learning is employed between \(\mathbf{E}^S_k\) and \(\mathbf{queue^T}\) to facilitate knowledge transfer from the teacher network \(f^{(T)}\) to the student network \(f^{(S)}\). In accordance with the MoMA framework, a classifier is devised to learn the action class.
This study uses the EfficientNet-B0 \cite{tan2019efficientnet} as the backbone for all the Image-MoMA-based models. 

\subsubsection{Objective function.}
The attention contrastive distillation is optimized using the Noise-Contrastive Estimation loss $\mathcal{L}_{\textnormal{\small Info}NCE}$ that is proposed in  \cite{oord2018representation}. The $\mathcal{L}_{\textnormal{\small Info}NCE}$ for an image $x_i$, with features feature \(\ e^S \in \mathbf{E}^T_k\) from student model and feature \( e^{T} \in \mathbf{E}^T_k\) from teacher model, is formulated as:
\begin{equation}
\mathcal{L}_{\textnormal{\small Info}NCE} =  -\log\frac{\exp(e_i^S \cdot e_i^{T} /\tau)}{ \sum_{i,j=0}^{L_N}\exp(e_i^S \cdot e_j^{queue^T}/\tau)},
\end{equation}
where $\tau$ is a temperature hyper-parameter ($\tau=0.07$). By minimizing $\textnormal{\small Info}NCE$, we maximize the mutual information between the positive pairs, i.e., \(\mathbf{E}^T_k\) and \(\mathbf{E}^S_k\), and minimize the similarity between $e^{S}$ and negative samples from $\mathbf{queue^T}$. 

For supervised training in the action recognition task, we utilize pairs \((\mathbf{F}_i, \mathbf{y}_i)\) provided by the organizers. In the context of action anticipation, we straightforwardly construct pairs \((\mathbf{F}_i, \mathbf{y}_{i+1})\). The supervised tasks are optimized using cross-entropy loss $\mathcal{L}_{CE}$. 

The overall objective function is as follows:
\begin{equation}
\mathcal{L} = \alpha\mathcal{L}_{CE} + \beta \mathcal{L}_{\textnormal{\small Info}NCE},
\end{equation}
where $\alpha = \beta = 1$.

\subsubsection{MoMA pre-trained weight.}
First, we utilized ImageNet pre-trained weights to initialize both the student and teacher networks. Secondly, drawing insights from MoMA, we also employed pre-trained weights from the action recognition task to transfer knowledge to both action recognition and action anticipation. The final results for the two tasks were obtained by aggregating outcomes from two sources: MoMA initialized with ImageNet pre-trained weights (ImageNet-MoMA) and MoMA initialized with action recognition task pre-trained weights (ActionRec-MoMA). Due to time constraints, the AGD learning for action recognition was based on an aggregation of four runs with ImageNet-MoMA and two runs with ActionRec-MoMA, while action anticipation utilized two runs for each.

\subsubsection{Comparative experiment.}
We compare our method with Video-ViT pre-trained on masked video distillation \cite{wang2023masked}. Similar to our image-based method, we randomly sampled 16 frames and fit them into Video-ViT.  Video-ViT includes a 3D-CNN layer patch embedding and ViT-transform encoder, and one classification head in single-task and two classifier heads in multi-task classification. This study uses the ViT-small \cite{dosovitskiy2020image} as the backbone for all the Video-ViT models. 
We also compare our ADG learning with single-task and multi-task learning on the MoMA framework.

\subsection{Track 2 - Task 1: Visual Question Answering (VQA)}

\begin{figure}[!t]
\centerline{\includegraphics[width=1\linewidth]{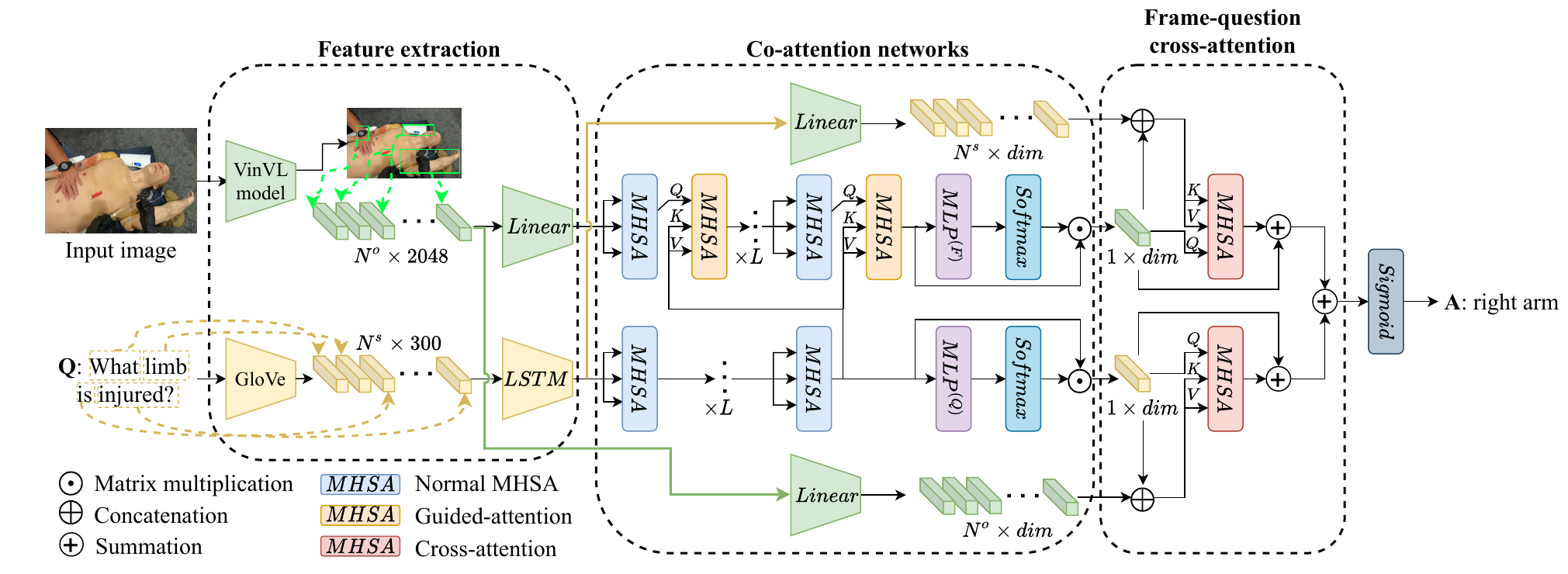}}
\caption{Illustration of our model for VQA task.}
\label{vqa-model}
\end{figure}

In this section, we first introduce the definition of the problem, and then we present clearly our proposed solution for this task. Figure \ref{vqa-model} illustrates the overview of our solution.

\subsubsection{Problem definition.} For the VQA task of the T3 challenge, the goal is to answer the questions provided per frame in videos. Hence, we can formulate the problem as: given a video \(\mathbf{V}\) including \(N\) frames, and a set of \(\mathbf{Q}^{(i)} = \Bigl\{q^{(i)}_k\Bigl\}^{N^q}_{k=1}\) including \(N^q\) questions for each frame \(I_i\), the target is predicting the answers for those questions:

\begin{equation}
\renewcommand{\arraystretch}{1.3}
    \begin{array}{ll}
    \hat{a}^{(i)}_k = P\Big(a^{(i)}_k|I_i, q^{(i)}_k\Big) = \texttt{VQA\_model}\Big(I_i, q^{(i)}_k\Big), \\
    \mathbf{\hat A}^{(i)} = \Bigl\{\hat a^{(i)}_k\Bigl\}^{N^q}_{k=1}, \\
    \end{array}
\end{equation}
where \(\hat{a}^{(i)}_k\) is the predicted answer for k-th question of \(i^{th}\) frame \(v_i\), and then we obtain the set of answers \(\mathbf{\hat A}^{(i)}\) for set of questions \(\mathbf{Q}^{(i)}\). Finally, we can obtain the full answers \(\mathbf{\hat A}\) for a video \(\mathbf{I}\).

In this section, we present our approach for building the VQA model (\(\texttt{VQA\_model}(\cdot)\)) for this task.

\subsubsection{Video-to-frame processing.} The organizer supplied the dataset, which comprises 125 training videos and 71 testing videos. Each video has a frame length ranging from 809 to 19583, and for each frame, there are 8 to 15 associated questions, with each question having a single answer. In total, there are 15 types of questions and 16 types of answers, with varying numbers of possible answers for each question type. Although question-answer annotations are provided at the frame level, we observed that a sequence of multiple frames can share the same questions and answers. Given the high number of frames per video, utilizing all of them would be time-consuming. Consequently, for each video, we opted to sample one frame per every 15 frames, resulting in a significant reduction in the number of frames used for training while still maintaining answer accuracy. It is important to note that during inference, we are required to predict answers on a per-frame basis. To achieve this, we partition a video into \(N^L\) sequential frames, with each sequence comprising 15 frames. Notably, the final frame within each sequence serves as a representative for all preceding 14 frames, shouldering the responsibility of answering all questions pertaining to those 14 frames as well as itself.

\subsubsection{Feature extraction.}
Feature extraction is always a crucial aspect of good performance in vision-language (VL) tasks \cite{vaswani2017attention}. Two types of features are used for VL tasks: grid features and object features. Grid features capture the global context and can be considered the same as features extracted from CNNs, while object features, including feature vectors for specific detected objects on an image, can be regarded as local features. In the context of Medical Visual Question Answering (VQA), particularly concerning life-saving intervention (LSI) procedures, it is crucial to emphasize the significance of object recognition, including hands, tools, and other relevant items. This recognition is essential to ensure the provision of accurate and potential answers. Hence, we choose the VinVL pre-trained model \cite{zhang2021vinvl}. This model is based on Faster R-CNN \cite{girshick2015fast} but trained on large-scale datasets. For feature extraction in VQA of the T3 challenge, we utilized this pre-trained model. For a video \(\mathbf{V}\) including \(N\) frames, for each frame, we obtain the set of \(N^o \times 2048\) feature vectors, where \(N^o\) is the number of objects detected by the model.

% by merging OpenImages \cite{kuznetsova2020open}, Objects365 \cite{shao2019objects365}, COCO \cite{lin2014microsoft} and VG \cite{krishna2017visual}. The backbone network is ResNeXt-152 \cite{xie2017aggregated}. 

\subsubsection{Deep modular co-attention networks (MCAN).}
For the VQA model, we adopted the \textbf{D}eep \textbf{M}odular \textbf{C}o-\textbf{a}ttention \textbf{N}etworks (MCAN) \cite{yu2019deep} as the baseline. This model was the crucial milestone for the VQA problem. Given the set of object feature vectors for \(i^{th}\) frame \(\mathcal{F} \in \mathbb{R}^{N^o \times 2048}\), and the text question \(q\), first, the GloVe pre-trained word embeddings \cite{pennington2014glove} are leveraged to provide embedding vectors for question words. Then, from \(q\), the embedding vectors for questions \( \mathcal{Q} \in \mathbb{R}^{N^s \times 300} \) are obtained, where \(N^s\) is the number of words/tokens of a question. Then, an LSTM layer is used to process \(Q\) and map the original GloVe dimension to the hidden dimension (\(dim\)). There are two model flows \(f^{(F)}\) and \(f^{(Q)}\) which are based on multi-head self-attention (\texttt{MHSA}) \cite{vaswani2017attention} to process \(\mathcal{F}\) and \(\mathcal{Q}\). Both \(f^{(F)}\) and \(f^{(Q)}\) have a stack of \(L\) \texttt{MHSA} layers. However, in \(f^{(Q)}\), there is a special type of \texttt{MHSA} called guided-attention followed by each normal self-attention. \(i^{th}\) guided-attention considered \(\mathcal{F}_{l}\) as query and \(\mathcal{Q}_L\) as key and value, where \(\mathcal{F}_{l}\) is the output of \(i^{th}\) normal \texttt{MHSA}, and \(\mathcal{Q}_L\) is the final output from \(f^{(Q)}\). In this manner, the model learns to prioritize object features that are relevant to the context of the question. After processed by \(f^{(F)}\) and \(f^{(Q)}\), \(\mathcal{F}_L \in \mathbb{R}^{N^o \times dim}\) and \(\mathcal{Q}_L \in \mathbb{R}^{N^s \times dim}\) are obtained. Then, a dimension reduction design is applied, which can be formulated as follows:

\begin{equation}
\renewcommand{\arraystretch}{1.3}
\begin{array}{ll}
    \alpha^{q} = \texttt{softmax}\big(\texttt{MLP}^{(Q)}(\mathcal{Q}_{L})\big), \texttt{  }
    \alpha^{f} = \texttt{softmax}\big(\texttt{MLP}^{(F)}(\mathcal{F}_{L})\big), \\
    {\tilde \mathcal{Q}} = {\alpha^{q}}^\mathsf{T} \odot \mathcal{Q}_{L}, \texttt{  } {\tilde \mathcal{F}} = {\alpha^{f}}^\mathsf{T} \odot \mathcal{F}_{L},
\end{array}
\end{equation}
where \(\texttt{MLP}^{(Q)}\) and \(\texttt{MLP}^{(F)}\) are MLP layers to reduce the dimension from \(dim\) to \(1\). Then, \(\texttt{softmax}(\cdot)\) is used to compute the weights per object or token in \(\mathcal{F}^{L}\) or \(\mathcal{Q}^{L}\), i.e., \(\alpha^{q}\) and \(\alpha^{f}\). Finally, matrix multiplication \(\odot\) is then performed to multiply \(\alpha^{q}\) with \(\mathcal{Q}^{L}\), and \(\alpha^{f}\) with \(\mathcal{F}^{L}\), to obtained \(\tilde \mathcal{Q} \in \mathbb{R}^{dim}\) and \(\tilde \mathcal{F} \in \mathbb{R}^{dim}\). Then, the summation of \(\tilde \mathcal{Q}\) and \(\tilde \mathcal{F}\) is performed to obtain fused features \(\mathcal{Z}\), and a \(\texttt{Sigmoid}(\cdot)\) is applied to map the fused \(dim\)-dimensional features \(\mathcal{Z}\) to \(N\)-dimensional logits, where \(N\) is the number of all possible answers in training set. Then, the Binary Cross-entropy (BCE) loss function is used to train the N-way answer classification problem.

\subsubsection{Frame-question cross-attention (FQCA).}
As mentioned in the previous section, the dimension reduction design produced question features $\tilde{\mathcal{Q}}$ and object features $\tilde{\mathcal{F}}$ now are only single feature vectors and then are used to predict the answer. Herein, we design the cross-attention at the head of $f^{(F)}$ and $f^{(Q)}$, that helps the single question feature vector $\tilde{\mathcal{Q}} \in \mathbb{R}^{dim}$ can be aware of the set of object features in a frame $\mathcal{F} \in \mathbb{R}^{N^o \times 2048}$, and the same with $\tilde{\mathcal{F}} \in \mathbb{R}^{dim}$ and $\mathcal{Q} \in \mathbb{R}^{N^s \times dim}$. Note that, $\mathcal{F}$ and $\mathcal{Q}$ mentioned here are object features and question features before being passed to $f^{(F)}$ and $f^{(Q)}$. We argue that pre-trained VinVL and GloVe are strong enough to produce the raw representations for a frame and a question, so it makes sense to let $\tilde{\mathcal{F}}$ and $\tilde{\mathcal{Q}}$ interpolate information from those raw features.

We are inspired by the cross-attention mechanism \cite{chen2021crossvit}, which allows only one token to be performed self-attention to a sequence of tokens, to design the frame-question cross-attention. In terms of \(\tilde \mathcal{Q}\)\ and \(\mathcal{F}^L\), the cross-attention can be formulated as:

\begin{equation}
\renewcommand{\arraystretch}{1.3}
\begin{array}{ll}
\mathcal{F}_L' = \texttt{Linear\_layer}(\mathcal{F}_L), \\
\tilde \mathcal{Q}' = \texttt{Concatenate}(\tilde \mathcal{Q}, \mathcal{F}_L'), \\
\tilde\mathcal{Q}^{''} = \tilde \mathcal{Q} + \texttt{FC}^{(1)}\Big(\texttt{FC}^{(0)}(\tilde\mathcal{Q}) + \texttt{LN}\big(\texttt{MHSA}(\tilde \mathcal{Q}, \tilde \mathcal{Q}', \tilde \mathcal{Q}')\big)\Big),
\end{array}
\label{eq:cross-att}
\end{equation}
where \(\texttt{Linear\_layer}(\cdot)\) is a skip-projection to reduce the original dimension of \(\mathcal{F}_L\) to match \(\tilde{\mathcal{Q}}'\); \(\tilde{\mathcal{Q}}' \in \mathbb{R}^{(N^o + 1) \times dim}\) is the concatenation of \(\tilde{\mathcal{Q}}\) and \(\mathcal{F}_L'\); \(\texttt{MHSA}(\cdot)\) is the multi-head self-attention that takes \(\tilde{\mathcal{Q}}\) as query and \(\tilde{\mathcal{Q}}'\) as key and value; \(\texttt{FC}^{(0)}\) and \(\texttt{FC}^{(1)}\) are both alignment projection layers; \(\texttt{LN}(\cdot)\) is the normalization layer.

After Eq. \ref{eq:cross-att}, \(\tilde\mathcal{Q}^{''} \in \mathbb{R}^{dim}\) is still a \(dim\)-dimensional feature vector, but aware of the context of sequence of object feature vectors in \(\mathcal{F}_L\), and the awareness is compressed into \(dim\) dimensions. We also perform the same Eq. \ref{eq:cross-att} in terms of \(\tilde\mathcal{F}\) and \(\mathcal{Q}_L\), to obtain \(\tilde\mathcal{F}^{''} \in \mathbb{R}^{dim}\). After that, as same as MCAN \cite{yu2019deep}, the summation of \(\tilde\mathcal{Q}^{''}\) and \(\tilde\mathcal{F}^{''}\) is computed, followed by a \(\texttt{Sigmoid}(\cdot)\) to product \(N\)-dimensional logits.

\section{Results}\label{sec:results}

\subsection{Track 1 - Task 1 \& 2: Action Recognition \& Anticipation}

We report our performance for action recognition and anticipation tasks in Table \ref{tab:result-reg} and  \ref{tab:result-ant}, respectively. As previously mentioned, action recognition and anticipation encompass two sub-tasks: verb and noun predictions. We face the choice of either training two separate models for these tasks or pursuing an end-to-end multi-task learning approach. Since the end goal of the two tasks is to predict the action that requires the model to predict correctly both verb and noun at the same time, our ADG showed robust performance compared to both single-task and multi-task learning.

\subsubsection{Action recognition.} Initially, we implemented the Video-ViT framework proposed by \cite{wang2023masked} and trained two distinct networks for each task. However, the results exhibited subpar performance, with only \(5.74\%\) accuracy in terms of \(\mathbf{Acc}_{action}\), despite relatively better performance per task, achieving \(23.17\%\) and \(20.20\%\) accuracy for \(\mathbf{Acc}_{verb}\) and \(\mathbf{Acc}_{noun}\), respectively. This led us to the realization that excelling in individual tasks may not suffice. Consequently, we devised a multi-task learning strategy for the Video-ViT model, resulting in improved overall performance, yielding an \(\mathbf{Acc}_{action}\) of \(8.32\%\). We applied the same multi-learning strategy to the MoMA setting proposed by \cite{vuong2023moma}, leading to a performance boost compared to the multi-task Video-ViT model, with an increase of \(+5.74\%\) in \(\mathbf{Acc}_{action}\). Finally, our AGC learning improved to  \(\mathbf{Acc}_{action}\) accuracy to \(15.45\%\) despite both \(\mathbf{Acc}_{verb}\) and \(\mathbf{Acc}_{noun}\) are inferior compare to the multi-task Image-MoMA. The results show the importance of simple changes in task design that target the final goal of action recognition and are able to improve performance without consuming extra computational resources. Our AGD learning approach utilizes the fewest parameters when compared to both single-task learning and multi-task learning.

\begin{table}[]
\centering
\setlength{\tabcolsep}{2pt} % Default value: 6pt
\caption{Results on the action recognition task. \textbf{Bold} and \underline{underline} texts highlight best and second-best results, respectively.}
\resizebox{1\textwidth}{!}{\begin{tabular}{lcccc}
\hline
        \textbf{Task design} & \textbf{Network} & \(\mathbf{Acc_{action} (\%)}\)  & \(\mathbf{Acc_{verb} (\%)}\) & \(\mathbf{Acc_{noun} (\%)}\) \\ \hline
        2 single tasks & Video-ViT & 5.74 & \textbf{23.17} & 20.20 \\ 
        1 multi-task & Video-ViT & 8.32 & 21.39 & 18.22 \\ 
        1 multi-task & Image-MoMA & 14.06 & \underline{22.57} & \textbf{24.95} \\ \hline
        1 AGD-task  (Ours) & Image-MoMA & \textbf{15.45} & 21.98 & \underline{24.55} \\ \hline
        
        % 2 single tasks & Video-ViT & 5.74 & 23.17 & 20.20 \\
        % 1 multi-task & Video-ViT & 8.32 & 21.39 & 18.22 \\
        % 1 multi-task & Image-MoMA & 8.71 & 15.05 & 20.20 \\ 
        % 1 combine task & Image-MoMA & 14.06 & 19.41 & 21.98 \\ \hline
        % 1 multi-task$^*$ & Image-MoMA (Ours) & \textbf{14.65} & \textbf{24.36} & \textbf{26.73} \\ \hline
        % \multicolumn{4}{l}{\small * mean voting of 9 runs.} \\

\end{tabular}}
\label{tab:result-reg}
\end{table}
\subsubsection{Action anticipation.}
Our Track 1 - Task 2 Action anticipation results are shown in table \ref{tab:result-ant}
Similar to Track 1 - Task 1 Action recognition, in Track 1 - Task 2 Action anticipation, the single-task Video-ViT framework \cite{wang2023masked} also obtains a relatively good accuracy on verb and noun classification, but only \(4.21\%\) \(\mathbf{Acc}_{action}\) accuracy. Multi-task Video-ViT slightly improves the  \(\mathbf{Acc}_{action}\) accuracy to \(5.89 \%\). Since the action anticipation task is more challenging than the action recognition, training the Video-ViT framework on a small amount of training data is even harder. Image-based method with contrastive learning proposed by \cite{vuong2023moma} improves the multi-task performance to  \(10.53\%\) \(\mathbf{Acc}_{action}\) accuracy. Finally, our AGC learning improved to  \(\mathbf{Acc}_{action}\) accuracy to \(12.84\%\).

\begin{table}[]
\centering
\setlength{\tabcolsep}{2pt} % Default value: 6pt

\caption{Results on the action anticipation task. \textbf{Bold} and \underline{underline} texts highlight best and second-best results, respectively.}
\resizebox{1\textwidth}{!}{\begin{tabular}{lcccc}
\hline
        \textbf{Task design} & \textbf{Network} & \(\mathbf{Acc_{action} (\%)}\)  & \(\mathbf{Acc_{verb} (\%)}\) & \(\mathbf{Acc_{noun} (\%)}\) \\ \hline
        2 single tasks  & Video-ViT & 4.21 & 16.84 & 17.68 \\ 
        1 multi-task & Video-ViT & 5.89 & 17.47 & 18.53 \\ 
        1 multi-task & Image-MoMA  & \underline{10.53} & \textbf{19.58} & \underline{19.58} \\ \hline
        1 AGD-task  (Ours) & Image-MoMA  & \textbf{12.84} & \underline{18.32} & \textbf{21.05} \\ \hline
\end{tabular}}
\label{tab:result-ant}
\end{table}

\subsection{Track 2 - Task 1: Visual Question Answering (VQA)}

While the organizer provided 125 official training videos, we were able to use only 52 of them due to issues with frame extraction from some videos. These 52 videos were split into training and testing sets in an 8:2 ratio. Subsequently, the training set was employed to train the MCAN model using VinVL object features, and validation was performed on the validation set. The highest-performing model on the validation samples was selected for submission on the official testing set, including 71 videos.

In terms of evaluation metrics, we utilized the BLEU score \cite{papineni2002bleu} and Accuracy. The BLEU score is employed to measure the similarity between two sequences in n-gram words, a common practice in vision-language tasks. Accuracy, on the other hand, quantifies the total number of correct answers. It's worth noting that the BLEU score is only reported for our validation set (\(\mathbf{B@1_{val}}\) and \(\mathbf{B@4_{val}}\)), as the submission system did not provide BLEU scores for the official testing set. 

Two variants of the MCAN model are available: a \texttt{small} version with a hidden dimension of 512, and a \texttt{large} version with a hidden dimension of 1024. Higher dimensions enable the model to capture more information but may result in reduced computational efficiency. Our performance evaluation includes results for both the small and large versions of the MCAN model.

\begin{table}[]
\centering
\caption{Results of three experiment settings on the official test set for VQA task. \textbf{Bold} and \underline{underline} texts highlight best and second-best results, respectively}
\resizebox{1\textwidth}{!}{\begin{tabular}{lcccc}
\hline
\textbf{Method} & \(\mathbf{B@1_{val}}\) & \(\mathbf{B@4_{val}}\) & \(\mathbf{Acc_{val} (\%)}\) & \(\mathbf{Acc_{test} (\%)}\) \\
\hline
MCAN-\texttt{small} w/ VinVL & \underline{89.31} & \underline{44.61} & \underline{88.02} & 70.60    \\
MCAN-\texttt{large} w/ VinVL & 85.88 & 43.18 & \textbf{88.37} & \underline{72.90}    \\
\hline
MCAN-\texttt{large} w/ VinVL w/ \texttt{FQCA} (ours) & \textbf{92.62} & \textbf{44.87} & 87.88 & \textbf{74.35}    \\
\hline
\end{tabular}}
\label{tab:result-vqa}
\end{table}

As depicted in Table \ref{tab:result-vqa}, MCAN-\texttt{large} surpasses MCAN-\texttt{small} by \(0.35\%\) and \(2.3\%\) in terms of \(\mathbf{Acc_{val}}\) and \(\mathbf{Acc_{test}}\), respectively, albeit exhibiting a decrease in BLEU metrics of \(-3.43\) and \(-1.43\) for \(\mathbf{B@1_{val}}\) and \(\mathbf{B@4_{val}}\), respectively. Accuracy is reported on both validation set (\(\mathbf{Acc}_{val}\)) and official testing set (\(\mathbf{Acc}_{test}\)).

Upon the incorporation of our proposed FQCA mechanism, we achieved \(88.78\%\) and \(74.35\%\) in \(\mathbf{Acc_{val}}\) and \(\mathbf{Acc_{test}}\), respectively, marking the highest performance on the official testing set. However, there was a slight reduction in \(\mathbf{Acc_{val}}\) compared to MCAN-\texttt{large} and MCAN-\texttt{small}. Notably, when evaluating sequence-comparing metrics \(\mathbf{B@1_{val}}\) and \(\mathbf{B@4_{val}}\), our approach excelled, achieving \(92.62\%\) and \(44.87\%\), respectively.

\section{Conclusion}\label{sec:conclusion}

In summary, our participation in the T3 challenge covered action recognition, anticipation, and visual question answering. We achieved promising results in action recognition and participation by employing an action dictionary-guided design with a MoMA setting. For the VQA task, we utilized VinVL pre-trained models for feature extraction and introduced a frame-question cross-attention mechanism based on the MCAN model, leading to the best performance in our experiments.

\bibliographystyle{splncs04}
\bibliography{ref}

\begin{thebibliography}{10}
\providecommand{\url}[1]{\texttt{#1}}
\providecommand{\urlprefix}{URL }
\providecommand{\doi}[1]{https://doi.org/#1}

\bibitem{calway2015discovering}
Calway, A., Mayol-Cuevas, W., Damen, D., Haines, O., Leelasawassuk, T.: Discovering task relevant objects and their modes of interaction from multi-user egocentric video. In: BMVC. pp. 30--1 (2015)

\bibitem{chen2021crossvit}
Chen, C.F.R., Fan, Q., Panda, R.: Crossvit: Cross-attention multi-scale vision transformer for image classification. In: Proceedings of the IEEE/CVF international conference on computer vision. pp. 357--366 (2021)

\bibitem{damen2020epic}
Damen, D., Doughty, H., Farinella, G.M., Fidler, S., Furnari, A., Kazakos, E., Moltisanti, D., Munro, J., Perrett, T., Price, W., et~al.: The epic-kitchens dataset: Collection, challenges and baselines. IEEE Transactions on Pattern Analysis and Machine Intelligence  \textbf{43}(11),  4125--4141 (2020)

\bibitem{dosovitskiy2020image}
Dosovitskiy, A., Beyer, L., Kolesnikov, A., Weissenborn, D., Zhai, X., Unterthiner, T., Dehghani, M., Minderer, M., Heigold, G., Gelly, S., et~al.: An image is worth 16x16 words: Transformers for image recognition at scale. arXiv preprint arXiv:2010.11929  (2020)

\bibitem{girshick2015fast}
Girshick, R.: Fast r-cnn. In: Proceedings of the IEEE international conference on computer vision. pp. 1440--1448 (2015)

\bibitem{kirkpatrick2017damage}
Kirkpatrick, A.W., McKee, J.L., McBeth, P.B., Ball, C.G., LaPorta, A., Broderick, T., Leslie, T., King, D., Beatty, H.E.W., Keillor, J., et~al.: The damage control surgery in austere environments research group (dcsaerg): a dynamic program to facilitate real-time telementoring/telediagnosis to address exsanguination in extreme and austere environments. Journal of Trauma and Acute Care Surgery  \textbf{83}(1),  S156--S163 (2017)

\bibitem{li2018eye}
Li, Y., Liu, M., Rehg, J.M.: In the eye of beholder: Joint learning of gaze and actions in first person video. In: Proceedings of the European conference on computer vision (ECCV). pp. 619--635 (2018)

\bibitem{oord2018representation}
Oord, A.v.d., Li, Y., Vinyals, O.: Representation learning with contrastive predictive coding. arXiv preprint arXiv:1807.03748  (2018)

\bibitem{papineni2002bleu}
Papineni, K., Roukos, S., Ward, T., Zhu, W.J.: Bleu: a method for automatic evaluation of machine translation. In: Proceedings of the 40th annual meeting of the Association for Computational Linguistics. pp. 311--318 (2002)

\bibitem{pennington2014glove}
Pennington, J., Socher, R., Manning, C.D.: Glove: Global vectors for word representation. In: Proceedings of the 2014 conference on empirical methods in natural language processing (EMNLP). pp. 1532--1543 (2014)

\bibitem{pirsiavash2012detecting}
Pirsiavash, H., Ramanan, D.: Detecting activities of daily living in first-person camera views. In: 2012 IEEE conference on computer vision and pattern recognition. pp. 2847--2854. IEEE (2012)

\bibitem{tan2019efficientnet}
Tan, M., Le, Q.: Efficientnet: Rethinking model scaling for convolutional neural networks. In: International conference on machine learning. pp. 6105--6114. PMLR (2019)

\bibitem{vaswani2017attention}
Vaswani, A., Shazeer, N., Parmar, N., Uszkoreit, J., Jones, L., Gomez, A.N., Kaiser, {\L}., Polosukhin, I.: Attention is all you need. Advances in neural information processing systems  \textbf{30} (2017)

\bibitem{vuong2023moma}
Vuong, T.T.L., Kwak, J.T.: Moma: Momentum contrastive learning with multi-head attention-based knowledge distillation for histopathology image analysis. arXiv preprint arXiv:2308.16561  (2023)

\bibitem{wang2023masked}
Wang, R., Chen, D., Wu, Z., Chen, Y., Dai, X., Liu, M., Yuan, L., Jiang, Y.G.: Masked video distillation: Rethinking masked feature modeling for self-supervised video representation learning. In: Proceedings of the IEEE/CVF Conference on Computer Vision and Pattern Recognition. pp. 6312--6322 (2023)

\bibitem{yu2019deep}
Yu, Z., Yu, J., Cui, Y., Tao, D., Tian, Q.: Deep modular co-attention networks for visual question answering. In: Proceedings of the IEEE/CVF conference on computer vision and pattern recognition. pp. 6281--6290 (2019)

\bibitem{zhang2021vinvl}
Zhang, P., Li, X., Hu, X., Yang, J., Zhang, L., Wang, L., Choi, Y., Gao, J.: Vinvl: Revisiting visual representations in vision-language models. In: Proceedings of the IEEE/CVF conference on computer vision and pattern recognition. pp. 5579--5588 (2021)

\end{thebibliography}
\end{document}